\documentclass[10pt,twocolumn,letterpaper]{article}

\usepackage[pagenumbers]{cvpr} 

\usepackage[dvipsnames]{xcolor}
\usepackage{graphicx}
\usepackage{amsmath}
\usepackage{amssymb}
\usepackage{booktabs}
\usepackage{array}
\usepackage{float}
\usepackage{dblfloatfix}

\newcommand{\methodname}{\textbf{DeforHMR}}

\definecolor{cvprblue}{rgb}{0.21,0.49,0.74}
\usepackage[pagebackref,breaklinks,colorlinks,citecolor=cvprblue]{hyperref}

\def\paperID{409} 
\def\confName{3DV\xspace}
\def\confYear{2025\xspace}

\title{DeforHMR: Vision Transformer with Deformable Cross-Attention for 3D Human Mesh Recovery  }

\author{\textbf{Jaewoo Heo}\textsuperscript{1}\\
Stanford University\\
{\tt\small jeffheo@stanford.edu}
\and
\textbf{George Hu}\textsuperscript{1}\\\
Stanford University\\
{\tt\small gehu@stanford.edu}
\and
\textbf{Zeyu Wang} \\\
Stanford University\\
{\tt\small wangzeyu@stanford.edu}
\and
\textbf{Serena Yeung-Levy}\\\
Stanford University\\
{\tt\small syyeung@stanford.edu}
}

\begin{document}
\twocolumn[{
\maketitle
\begin{center}
{\footnotesize \textsuperscript{1}Equal contribution}
    \captionsetup{type=figure}
    \includegraphics[width=\textwidth]{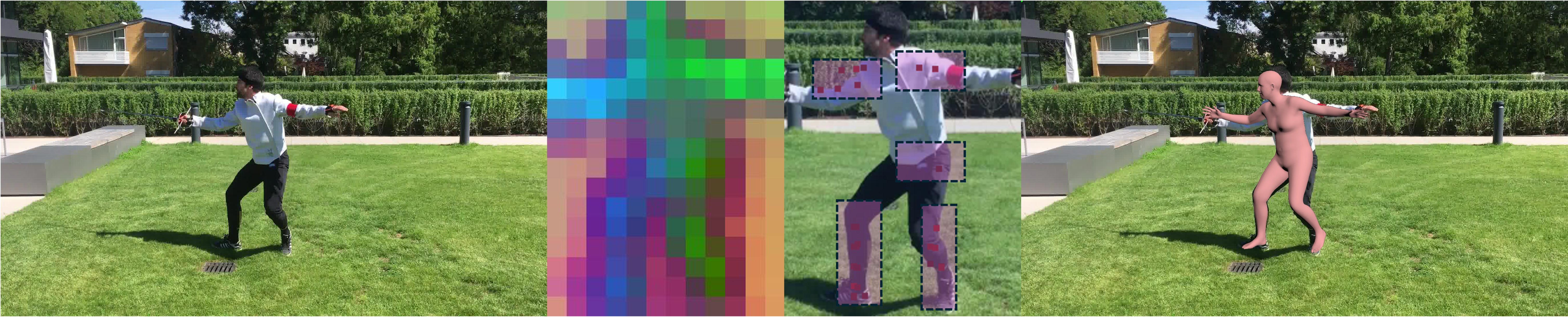}
    \caption{We present \methodname, a single-image, regression-based methodology for HMR. \methodname\ uses a vision transformer (ViT) encoder to derive spatial features from the input image and a deformable cross-attention transformer decoder to learn meaningful spatial relations from the features, enabling the ability to recover accurate 3D human body meshes. \textbf{Left to right:} \textbf{(1)} An image in the "outdoors\_fencing\_01" class of 3DPW \cite{vonMarcard2018}. \textbf{(2)} Spatial feature of the image outputted by our ViT encoder. \textbf{(3)} Attention visualization for the first head of the first layer of our deformable attention transformer decoder. Red and maroon square dots are attention locations; we emphasize the boxed and highlighted areas. \textbf{(4)} \methodname's output mesh projected onto original image.}
    \label{fig:title_figure}
\end{center}
\vspace{0.75cm}
}]

\begin{abstract}
Human Mesh Recovery (HMR) is an important yet challenging problem with applications across various domains including motion capture, augmented reality, and biomechanics. Accurately predicting human pose parameters from a single image remains a challenging 3D computer vision task. In this work, we introduce \methodname, a novel regression-based monocular HMR framework designed to enhance the prediction of human pose parameters using deformable attention transformers. \methodname\ leverages a novel query-agnostic deformable cross-attention mechanism within the transformer decoder to effectively regress the visual features extracted from a frozen pretrained vision transformer (ViT) encoder. The proposed deformable cross-attention mechanism allows the model to attend to relevant spatial features more flexibly and in a data-dependent manner. Equipped with a transformer decoder capable of spatially-nuanced attention, \methodname\ achieves state-of-the-art performance for single-frame regression-based methods on the widely used 3D HMR benchmarks 3DPW and RICH. By pushing the boundary on the field of 3D human mesh recovery through deformable attention, we introduce an new, effective paradigm for decoding local spatial information from large pretrained vision encoders in computer vision.

\end{abstract}
    
\section{Introduction}
\label{sec:intro}

Motion capture (MoCap) technology has applications in numerous fields such as film, gaming, AR/VR, as well as sports medicine by providing a tool to capture and analyze human pose in 3D. Traditional marker-based MoCap systems utilizing multi-view cameras and marker suits recover highly accurate human pose but suffer from poor accessibility due to the high cost of setting up the adequate laboratory environment \cite{ma20243dhumanmeshestimation}. In contrast, a single camera with the correct algorithm can perform 3D Monocular Human Mesh Recovery (HMR), which recovers a mesh of a human body in 3D given an input image or video as a more accessible alternative using deep neural networks \cite{kanazawa2018endtoend}.

A common parametric approach to 3D HMR leverages the Skinned Multi-Person Linear (SMPL) \cite{loper2015smpl} representation model that regresses joint articulations (often referred to as the pose parameter) and a body shape parameter to generate accurate 3D human body meshes. Current challenges in HMR include occlusion situations \cite{khirodkar2022occludedhumanmeshrecovery} and the complexity and variability of human pose, but underlying these issues is simply insufficient spatial understanding in neural networks to output correct pose parameters \cite{li2022cliff}.  

More recently, advances in vision transformers \cite{dosovitskiy2021imageworth16x16words} have demonstrated versatility and overall impressive performance across a wide range of vision tasks and domains \cite{Khan_2022}, particularly in determining complex spatial relations \cite{goel2023humans}. In the field of object detection, deformable attention \cite{zhu2021deformable} \cite{xia2023datspatiallydynamicvision} has emerged as one promising solution for accurate, space-aware localization, and extending such an approach to HMR requires even greater focus on extracting precise positional semantics \cite{zhu2021deformable}. 

In parallel, issues of data generalization across diverse real world applications for vision models have been diminished by the release of large vision transformer models pretrained on self-supervision tasks on web-scale datasets \cite{oquab2024dinov2} \cite{bommasani2022opportunities}. The ability of these foundation models to generate meaningful features across all data spectra for downstream application has created a new effective learning paradigm, and more recently, works \cite{fu2024featup} have begun on improving the spatial resolution of these vision foundation model features for ever better results. For HMR, \cite{goel2023humans} has noted how pretraining with both masked auto-encoding \cite{he2021maskedautoencodersscalablevision} along with 2D keypoint prediction \cite{xu2022vitposesimplevisiontransformer} \cite{cao2019openposerealtimemultiperson2d}  has been essential to model convergence, and we build upon along this line of work, investigating how to most effectively decode the features from these large-scale pretrained models. 

Integrating the information derived from large, pretrained vision transformer \cite{xu2022vitposesimplevisiontransformer} features and deformable attention decoding, we present \methodname, a novel transformer-based HMR framework that significantly improves upon current methods in both accuracy and computation efficiency.

We believe \methodname\ offers significant benefit through the synergy between the semantically-meaningful spatial features from a pretrained vision transformer and the deformable attention mechanism; in deformable attention the reference and offset locations are floating point values in the feature map coordinate space, and bilinear interpolation is used to extract the relevant key and value information. Hence, the advantage of such deformable attention mechanism is derived mostly from data-dependent spatial flexibility, or the ability to dynamically shift attention to relevant spatial regions based on the characteristics of the input feature. We believe utilizing rich features from the transformer encoder would enable the spatial dynamism of the deformable attention mechanism to be influential by learning better spatial relations.

In summary, our contributions are twofold:
\begin{enumerate}
    \item We present \methodname, a regression-based monocular HMR framework that demonstrates SOTA performance on multiple well-known public 3D HMR datasets under the single-frame, regression-based setting.
    \item Inspired by \citep{xia2023datspatiallydynamicvision}, we propose a novel deformable cross-attention mechanism designed to be query-agnostic and spatially flexible.
\end{enumerate}

\section{Related Work}
\label{sec:related_work}

\subsection{Monocular HMR}
Early work in 3D HMR \cite{pavlakos2019expressivebodycapture3d}\cite{Bogo:ECCV:2016} revolved mainly around fitting the SMPL parametric body model to minimize the discrepancy between its reprojected joint locations and 2D keypoint predictions on the 2D image. End-to-end 3D human mesh recovery, not relying upon intermediate 2D keypoints or joints, was first proposed by Kanazawa et al. \cite{kanazawa2018endtoend}. This was achieved by leveraging novel deep learning advancements at the time and regressing the SMPL parameters along with a camera model to derive the 3D human meshes. Ever since then, various neural network-based HMR methodologies have been proposed. In \cite{li2022hybrikhybridanalyticalneuralinverse}, the authors aim to resolve the discrepancy between plausible human meshes and accurate 3D key-point predictions through a hybrid inverse kinematics solution involving twist-and-swing decomposition. Li et al. \cite{li2022cliff} proposes to mitigate the loss of global positional information after cropping the human body through utilizing more holistic features containing global location-aware information to ultimately regress the global orientation better, and, Goel et al. \cite{goel2023humans} would contribute to this through implementing a vision transformer architecture using a single query token fed into the decoder for SMPL parameter predictions. HMR-2.0 established a new competitive baseline on single human mesh recovery, and in particular, they show how their transformer network can encode and decode complex human pose due to their ability to better capture difficult spatial relations.

\subsubsection{Optimization-Based HMR}
In \cite{li2022hybrikhybridanalyticalneuralinverse}, Li et al. notes how an alternative approach to direct regression of the SMPL parameters is optimization-based HMR \cite{choutas2020monocularexpressivebodyregression}\cite{50649}\cite{xiang2018monoculartotalcaptureposing}\cite{Zioulis_2023}, which estimates the body pose and shape through an iterative fitting process. For instance, PLIKS \cite{shetty2023plikspseudolinearinversekinematic} fits a linearized formulation of the SMPL model on region-based 2D pixel-alignment, and ReFit \cite{Wang_2023_ICCV} iteratively projects 2D keypoints in order to effectively generate accurate meshes. However, optimization-based HMR at inference-time does not have any runtime guarantees and often struggles from large runtime due to the iterative refinement process, and thus can be difficult to integrate into real-time application settings.

\subsubsection{Temporal HMR}
As computational capacity has increased over recent years, the ability to use complete sequences of video frames for human mesh recovery has recently found success. Within this area of temporal HMR \cite{doersch2019sim2realtransferlearning3d}\cite{luo20203dhumanmotionestimation}\cite{choi2021staticfeaturestemporallyconsistent}\cite{9710851}, Kocabas et al. \cite{kocabas2020vibevideoinferencehuman} have proposed adversarial training with a temporal network architecture to learn to discriminate between real and fake pose sequences. Moreover, \cite{yuan2022glamrglobalocclusionawarehuman} proposes to mitigate the effects of occlusions in video sequences through infilling occluded humans with a deep generative motion infiller, and \cite{ye2023decouplinghumancameramotion} utilizes a temporal motion prior \cite{rempe2021humor3dhumanmotion} to effectively decouple the human motion and camera motion given a video sequence. More recently, Shin et al. \cite{shin2024whamreconstructingworldgroundedhumans} have incorporated motion encoding and SLAM \cite{teed2022droidslamdeepvisualslam} approximation, along with model scale in order to achieve obtain state of the art performance for multi-frame inputs.

While this line of work is promising for integrating complete video information in human mesh recovery, we restrict our focus to single-frame inputs so that our method generalizes to individual images.

\subsection{Deformable Attention}
The Deformable Transformer \cite{zhu2021deformable} architecture, first proposed in end-to-end object detection \cite{carion2020endtoendobjectdetectiontransformers}, has demonstrated comparable performance to other SOTA methods without needing any hand-designed components commonly used in object detection \cite{redmon2016yolo9000betterfasterstronger}\cite{girshick2014richfeaturehierarchiesaccurate}. The deformable attention module is designed for efficiency and complex relational parameterization, having the keys and values be sparsely sampled learned offsets from a reference location determined by a given query. Zhu et al. \cite{zhu2021deformable} show that this increases model training and inference speed while also incorporating inductive biases for precise spatial modeling beneficial for object detection.

Yoshiyasu \cite{yoshiyasu2023deformable} extends this notion of deformable attention to optimization-based non-parametric 3D HMR with the DeFormer architecture, using the joint and shape query tokens at each layer to generate reference points and offsets on multi-scale maps to be used in the attention computation. DeFormer works directly with positional information without the SMPL parameterization for dense mesh reconstruction, and it improves upon previous baselines of similar model size.

In \cite{xia2023datspatiallydynamicvision}, Xia et al., show how previous works for deformable attention, in fact, function more like a deformable convolution \cite{dai2017deformableconvolutionalnetworks} without attention interactions between all queries and all keys. They then propose the Deformable Attention Transformer (DAT), a vision transformer backbone using true deformable self-attention. DAT demonstrates its advantage of deformable self-attention for localization-based tasks such as object detection and instance segmentation, outperforming shifted window full self-attention methods \cite{liu2021swintransformerhierarchicalvision} on COCO \cite{lin2015microsoft}. In their analysis, they suggest that deformable attention consistently attends to more important and relevant areas of the image and feature map compared to full self-attention, confirming that the true deformable attention interactions between queries and keys result in realized performance and interpretable improvements.



\section{Methodology}
\label{sec:methodology}

In this section, we delve into the methodologies of each component of \methodname. More specifically, we discuss using a frozen ViT pretrained on pose estimation as a feature encoder and our novel deformable cross-attention mechanism. Lastly, we touch upon model training specifications.




\subsection{Generating Feature Maps} 

We use the ViT-Pose from Goel et al. \cite{goel2023humans} as our initial feature encoder. This is a ViT-H with patch size 16 and input size 256 by 192 that is pretrained with masked autoencoding on ImageNet and 2D pose estimation on COCO \citep{xu2022vitposesimplevisiontransformer}. We freeze all the weights during training and pass the input image through to generate the features maps. That is, given an input image $x \in \mathbb{R}^{H'\times W'\times 3}$ and a patch size of 16, we represent the spatial output tokens of the encoder $f$ as $f(x) \in \mathbb{R}^{H \times W \times C}$ for $H = H'/16$ and $W = W'/16$.


\subsection{Deformable Attention Decoder}
\subsubsection{SMPL Multi-query Transformer Decoder}

\begin{figure}[ht]  
    \centering
    \includegraphics[width=\columnwidth]{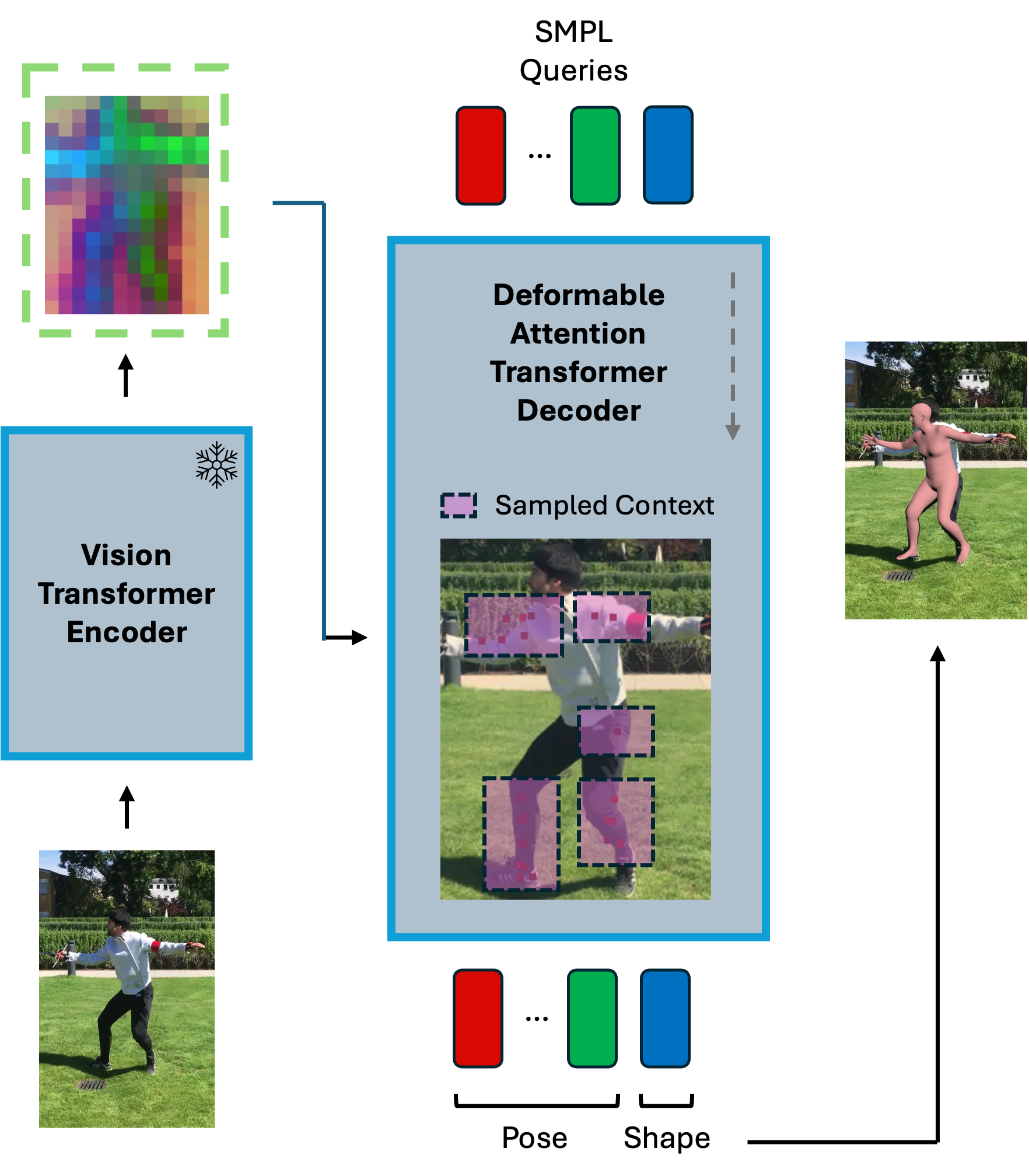}
    \caption{Full system architecture of \methodname. We dedicate a learnable query embedding for each of the 24 joint articulations and the body shape vector.}
    \label{fig:system}
\end{figure}



Our approach can be thought of as using query tokens for SMPL parameters. These are learnable tokens $t \in \mathbb{R}^{25 \times 1024}$, representing 24 pose tokens and 1 shape token, which further incorporate information from the image features through the decoder blocks.


Following the standard paradigm of the transformer decoder \cite{vaswani2023attentionneed}, each layer of our deformable cross-attention decoder is compromised of a self-attention, a deformable cross-attention, and a feed-forward network.
We further elaborate on our deformable cross-attention mechanism in the subsequent section.

After passing the queries through the decoder, we learn linear projections $W_{pose}$ and $W_{shape}$ to get the desired outputs of pose parameters $\theta \in \mathbb{R}^{24 \times 6}$, and shape parameters $\beta \in \mathbb{R}^{10}$. For the pose rotation angles in the SMPL parameters, we use the common 6D representation proposed by Zhou et al. (2020) \cite{zhou2020continuity} for a more continuous loss-landscape, converting to the actual pitch/roll/yaw and rotation matrix afterwards. Moreover, we use one round of iterative error feedback \cite{carreira2016human}, starting with the mean SMPL values from Humans 3.6M \cite{h36m_pami} to condition our predictions better. These are thus finally passed into the SMPL model to generate our 3D meshes.

\subsubsection{Deformable Cross-Attention}
\begin{figure*}[ht]  
    \centering     
    \includegraphics[width=\textwidth]{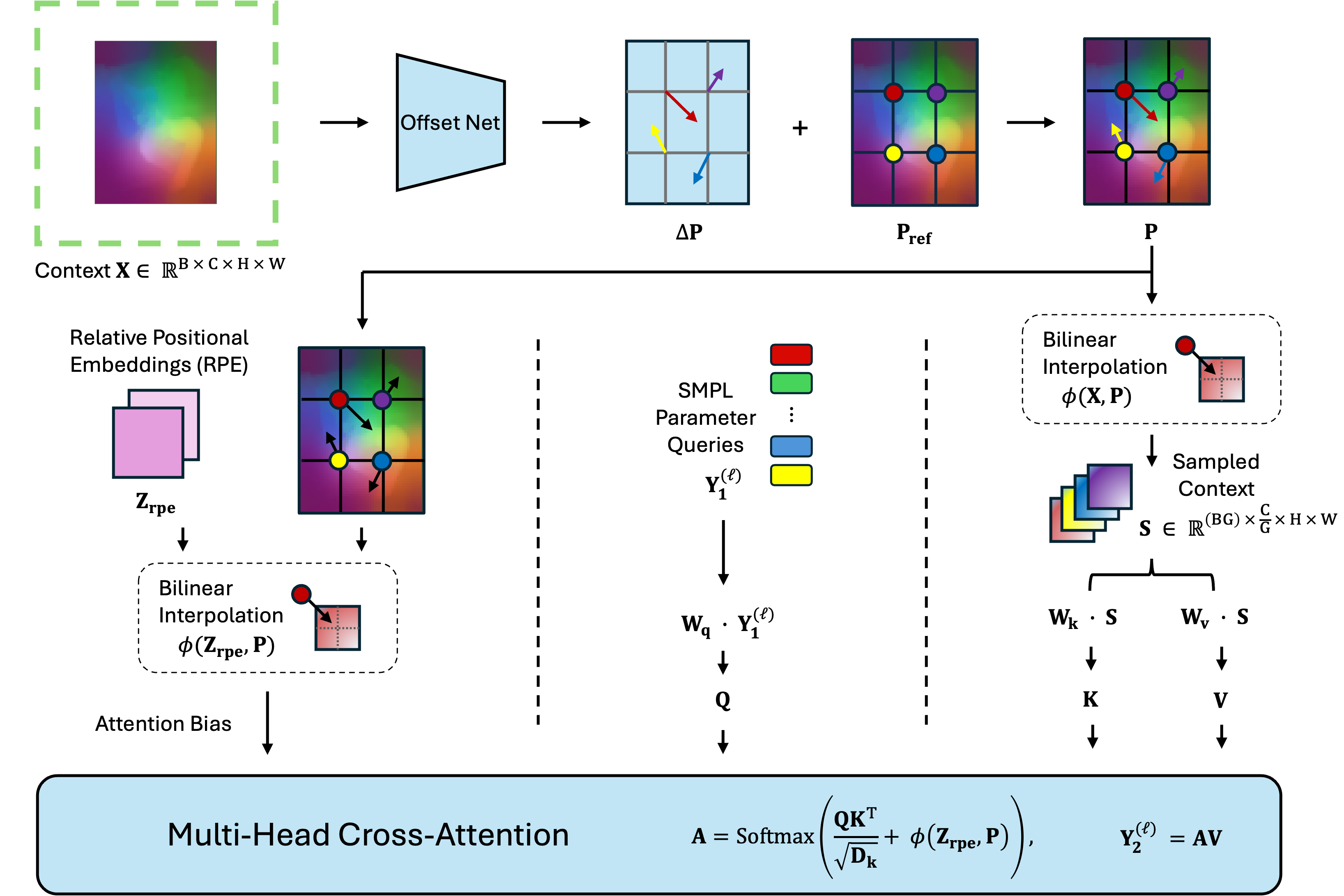}
    \caption{Our proposed deformable cross-attention module. The offset-generating convolutional neural network takes the spatial features from the transformer encoder to generate sampling position offsets $\Delta P$. These offsets are added to the grid reference positions $P_{ref}$ to get our final sampling positions $P$. These sampling positions are used to $1)$ sample the input context via bilinear interpolation, which is then projected to keys and values for attention computation, as well as $2)$ sample the relative positional embeddings (RPE) to get our attention bias. These are combined in the standard multi-head cross-attention formulation with relative position biases to generate the output.}
    \label{fig:system}
\end{figure*}

We propose a novel, query-agnostic deformable cross-attention mechanism designed to capture fine-grained spatial details as shown in \ref{fig:system}. The deformable aspect introduces learnable offsets that allow the attention to adaptively select key-value pairs from the feature map, as opposed to uniform attention to all spatial locations. Our method is inspired by the self-attention mechanism proposed in \cite{xia2023datspatiallydynamicvision}.

For some layer $\ell$, let the input tokens to this layer be $\mathbf{Y}^{(\ell-1)}\in\mathbb{R}^{B \times 25 \times D}$, for batch size $B$, the 25 SMPL tokens, and model dimension $D=1024$. The first part of our decoder block is multi-head self-attention with residual on the query tokens, so let the output from the self-attention be $\mathbf{Y}_1^{(\ell)} = \text{Self-Attn}(\mathbf{Y}^{(\ell-1)}) + \mathbf{Y}^{(\ell-1)}$ Let the spatial features from the encoder be $\mathbf{X} \in \mathbb{R}^{B \times H \times W \times C}$. We will refer to these spatial features as the \textit{context} for our decoder. 

We consider a base set of reference points $\mathbf{R} \in \mathbb{R}^{(BG) \times 2 \times H \times W}$, which are initialized as grid coordinates normalized to the range $[-1, 1]$. These reference points provide an initial uniform bias for the positions of the keys in the feature map:

\begin{equation}
\mathbf{R}_{ij} = \left(\frac{i}{H} \times 2 - 1, \frac{j}{W} \times 2 - 1\right),
\end{equation}

where $i$ and $j$ denote the indices over height and width, respectively.


We then compute $G$ unique raw offsets for each reference point by processing the context features through a series of grouped convolutions, non-linear activations (GELU) \cite{hendrycks2023gaussianerrorlinearunits}, and normalization layers:

\begin{equation}
\Delta \mathbf{P}' = \text{Conv2D}(\text{GELU}(\text{LayerNorm}(\text{Conv2D}(\mathbf{X})))).
\end{equation}

These raw offsets are then passed through hyperbolic tangent and scaled by $\frac{\lambda_o}{H_r}$, restricting the offset to within $\lambda_o$ multiplied by the grid spacing. Hence, the final offsets $\Delta \mathbf{P}$ are
\[\Delta \mathbf{P} = \frac{\lambda_o}{H_r}\tanh(\Delta\mathbf{P'})\]

The resulting offsets $\Delta \mathbf{P} \in \mathbb{R}^{(BG) \times 2 \times H \times W}$ indicate the amount by which each reference point in the grid of dimension $ H \times W$ should be shifted, allowing the model to focus on different parts of the feature map depending on the input context. This mechanism enables the attention to be more flexible and context-aware.



The final sampling positions are hence the sum of the reference point positions and offsets.

\begin{equation}
\mathbf{P} = \Delta \mathbf{P} + \mathbf{R}.
\end{equation}

With these positions, we sample the context $X$, employing bilinear interpolation to extract precise embedding values at these adjusted positions. The sampled context $\mathbf{S} \in \mathbb{R}^{(BG) \times \frac{C}{G} \times H \times W}$ can finally be projected into keys $\mathbf{K} = W_k \cdot \mathbf{S}$ and values $\mathbf{V} = W_v \cdot \mathbf{S}$. And likewise, we project the input tokens into the queries $\mathbf{Q} = W_q \cdot \mathbf{Y}_1^{(\ell)}$.


We calculate cross-attention scores by first taking the dot product of the queries $\mathbf{Q}$ with the keys $\mathbf{K}$, then summing this term with an attention bias term, which is computed through sampling a learnable relative positional embedding tensor $\mathbf{Z}_{rpe}$ via bilinear interpolation $\phi(\_ ,\_)$ using $\mathbf{P}$ as sampling location:

\begin{equation}
\mathbf{A} = \text{Softmax}\left(\frac{\mathbf{Q}  \mathbf{K}^T}{\sqrt{D_k}} + \phi(\mathbf{Z}_{rpe}, \mathbf{P}) \right)
\end{equation}

where $D_k$ is the dimension of the keys, used to scale the scores and stabilize training. 

More specifically, the relative positional embedding is a unique learned grid for each query position. This is similar to both DAT \cite{xia2023datspatiallydynamicvision} and the original relative position encoding \cite{shaw2018selfattentionrelativepositionrepresentations}; the keys have actual positions $P$, so it uses the grid formulation as in DAT, but the queries do not have positions, so we simply index and learn the relative position embedding separately for each query.

We lastly multiply the attention coefficients by the values and add the residual to get the cross attention output $\mathbf{Y}_2^{(\ell)} = \mathbf{A}\mathbf{V} + \mathbf{Y}_1^{(\ell)}$. Passing this through a 2-layer feed-forward network, we get the layer's final output $\mathbf{Y}^{(\ell)} = \text{FFN}(\mathbf{Y}_2^{(\ell)}) + \mathbf{Y}_2^{(\ell)}$


\subsubsection{Difference Between \methodname\ and Previous Deformable Cross-Attention}
We want to emphasize that our deformable cross-attention differs greatly from DeformableDETR\cite{zhu2021deformable}-style cross-attention proposed by previous works. Unlike typical deformable attention methods where offsets are conditioned on the queries, our model computes offsets directly from the context, meaning they are query-agnostic. This design choice is inspired by the Deformable Attention Transformer \cite{xia2023datspatiallydynamicvision} (DAT) paper; however, their focus on encoder architectures means their deformable self-attention models do not fully decouple relations between queries and key-values. By having query-agnostic cross-attention here, we can
ensure that the shifts in receptive fields and sampling clusters via deformable attention are consistent and coordinated across all queries, capturing global information more effectively.

\subsection{Training Details}

Following \cite{li2022cliff}, we train with reconstruction loss on the SMPL parameters, the 3D joint positions, the 3D mesh vertices, and the projected 2D joint positions, all using mean square error. The relative loss weight for SMPL parameters is $\lambda_{SMPL} = 1$, 2D and 3D joint positions is $\lambda_{joint} =5$, and mesh vertices $\lambda_{mesh} = 60$. For all training runs, we freeze the ViT-Pose to explore efficient decoding methods.

\begin{table*}[htb]
    \centering
    \begin{tabular}{lcccccccc}
        \toprule
        & \multicolumn{4}{c}{\textit{\textbf{3DPW}} \cite{vonMarcard2018}} & \multicolumn{4}{c}{\textit{\textbf{RICH}} \cite{huang2022capturinginferringdensefullbody}} \\
        \cmidrule(lr){2-5} \cmidrule(lr){6-9}
        \textbf{Method} & \textbf{PA-MPJPE} $\downarrow$ & \textbf{MPJPE} $\downarrow$ & \textbf{PVE} $\downarrow$ & & \textbf{PA-MPJPE} $\downarrow$ & \textbf{MPJPE} $\downarrow$ & \textbf{PVE} $\downarrow$ & \\
        \midrule
        ROMP \cite{sun2021monocularonestageregressionmultiple} & 47.3 & 76.7 & 93.4 &  & - & - & - &  \\
        PARE \cite{kocabas2021pareattentionregressor3d} & 46.5 & 74.5 & 88.6 &  & 60.7 & 109.2 & 123.5 &  \\
        CLIFF \cite{li2022cliff} & 43.0 & 69.0 & 81.2 &  & 56.6 & 102.6 & 115.0 &  \\
        HybrIK \cite{li2022hybrikhybridanalyticalneuralinverse} & 41.8 & 71.6 & 82.3 &  & 56.4 & 96.8 & 110.4 &  \\
        SA-HMR \cite{shen2023sahmr} & - & - & - &  & - & 93.9 & 103.0 &  \\
        HMR2.0\textsuperscript{$*$} \cite{goel2023humans} & 44.4 & 69.8 & 83.2 &  & \textbf{\underline{48.1}} & 96.0 & 110.9 &  \\
        
        \methodname\ & \textbf{\underline{38.3}} & \textbf{\underline{63.6}} & \textbf{\underline{75.2}} &  & 48.6 & \textbf{\underline{84.2}} & \textbf{\underline{94.5}} &  \\
        \bottomrule
    \end{tabular}
    \caption{Comparison of state-of-the-art models on 3DPW \cite{vonMarcard2018} and RICH \cite{huang2022capturinginferringdensefullbody} datasets. \methodname\ achieves superior HMR performance by a wide margin across all metrics on both datasets except PA-MPJPE on RICH that is comparable to that of HMR2.0 \cite{goel2023humans}. (*) represents the exclusion of 3DPW data during training.}
    \label{tab:main_results}
\end{table*}

We train all models on real world datasets, two with 3D SMPL ground truth derived from motion capture---3DPW \cite{vonMarcard2018} and MPI-INF-3DHP \cite{mono-3dhp2017}---and three psuedo-labeled from 2D pose ground truth using the CLIFF-annotator \cite{li2022cliff}: COCO \cite{lin2015microsoft}, MPII \cite{andriluka14cvpr}, and Humans3.6M \cite{h36m_pami}. We train for 100 epochs. The evaluation is performed on the test split of 3DPW and RICH, and we use mean per-joint position error (MPJPE), procrustes analysis MPJPE (PA-MPJPE), and per-vertex error (PVE), all in millimeters (mm) to determine how well the model recovers accurate human pose in 3D.

\section{Results}
\label{sec:results}

We compare various model architectures and approaches using our evaluation metrics in Table \ref{tab:main_results}. Note that since we are interested in single-frame inputs and inference in real-time applications, we exclude multi-frame temporal approaches and optimization-based approaches.

Upon comparing HMR evaluation metrics with several state-of-the-art regression-based HMR methodologies, we demonstrate that \methodname\ establishes a new state-of-the-art benchmark on both 3DPW \cite{vonMarcard2018} and RICH \cite{huang2022capturinginferringdensefullbody} datasets by a considerable margin. 


\begin{figure}[htb]
    \includegraphics[width=0.5\textwidth]{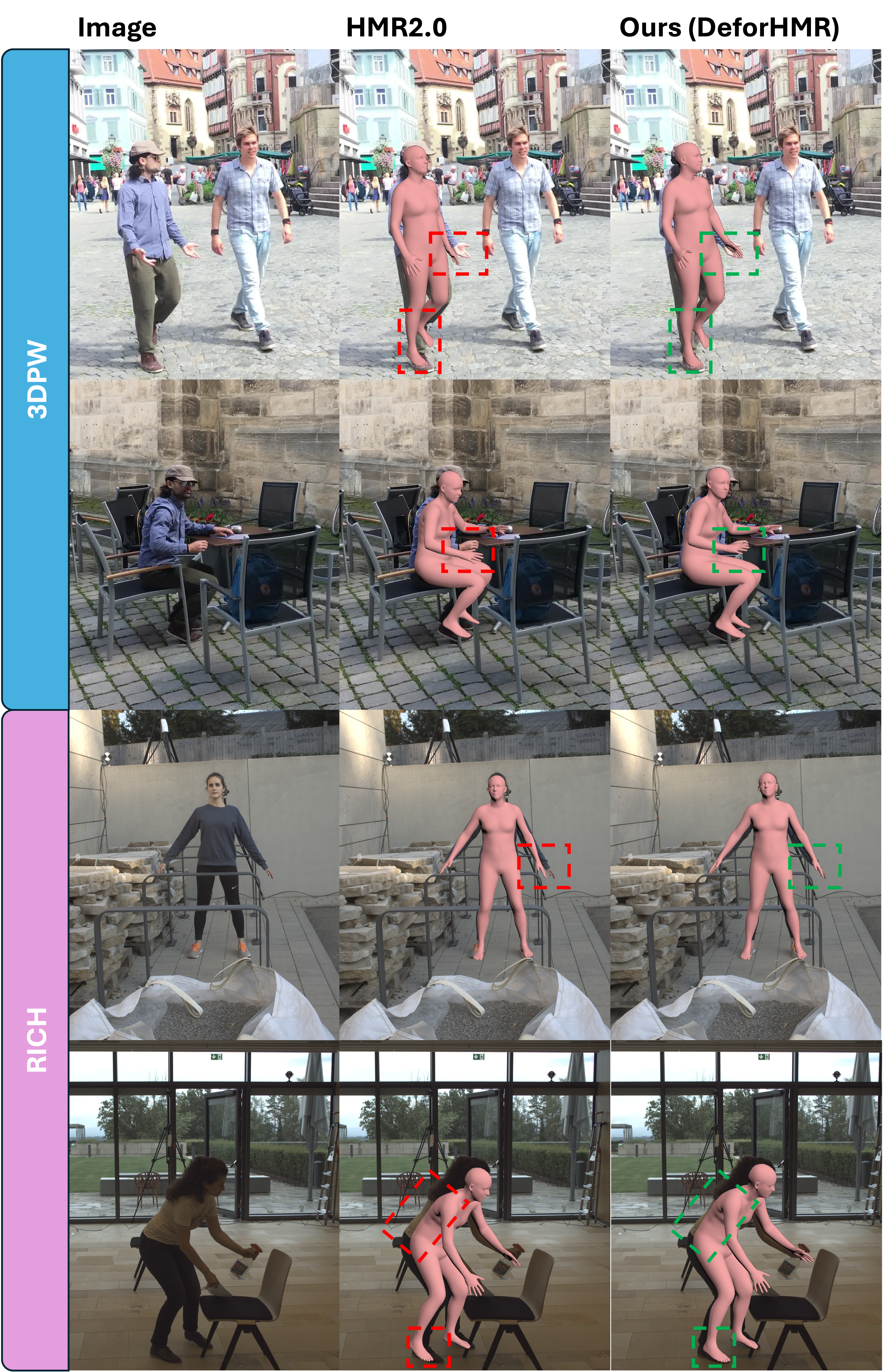}
    \caption{Qualitative results on test set. We visualize the original image and the predicted mesh projected onto the original image for both HMR2.0 \cite{goel2023humans} and \methodname. We highlight the inaccurate mesh regions outputted by HMR2.0 in red boxes and highlight the corresponding mesh region on \methodname's mesh output in green boxes. Upon visualizing HMR2.0 and our model's recovered meshes on four distinct scenarios across 3DPW \cite{vonMarcard2018} and RICH \cite{huang2022capturinginferringdensefullbody}, we observe \methodname's significant improvements on HMR2.0's ability to recover accurate 3D human meshes. While HMR2.0 shows inaccurate feet \& hand positions in all four rows as well as inaccurate orientation of the entire torso in the lowest row, \methodname\ consistently shows more accurate feet, hand, and torso positions.}
    \label{fig:qualitative}
\end{figure}


\subsection{Analysis}

Our HMR model exhibits a robust capability in capturing the general body pose and proportions of individuals across various scenarios, as seen in the visualizations on Figure \ref{fig:qualitative}. Upon rendering the recovered meshes on four distinct images from the test set of the 3DPW \cite{vonMarcard2018} and RICH\cite{huang2022capturinginferringdensefullbody} dataset, we confirm our model comprised of the ViT-Pose transformer encoder and the transformer decoder using deformable cross-attention generalizes well across various in-the-wild image examples. Our model demonstrates accurate, plausible, and realistic meshes for humans in various scenarios such as but not limited to executing a fencing motion, walking while conversing sideways, sitting at a table, crouching downwards, etc. In particular, compared to pre-existing HMR models, namely HMR2.0 \cite{goel2023humans}, we show strengths in accuracy of upper body articulation and orientation, as well as feet and hand position.

In table \ref{tab:query_ablation}, we decouple some of the main differences between HMR2.0\textsuperscript{$\dagger$} and  \methodname : multi-query decoder and deformable cross-attention. To do so, we evaluate all four combinations of $1)$ multi-query versus single-query, and $2)$ deformable cross-attention versus regular cross-attention on the test set of 3DPW \cite{vonMarcard2018}. The ablation results clearly indicate that both the use of multiple queries and the deformable cross-attention mechanism in \methodname\ contribute significantly to performance improvements across all three metrics. Specifically, models incorporating these components consistently achieve lower error rates, indicating the effectiveness of each design choice, and furthermore, the performance increase going from single to multi-query for deformable cross-attention is much larger than regular cross-attention (4.0mm PVE decrease versus 3.1). This suggests true synergy between the multi-query formulation and deformable cross-attention, enabling superior 3D HMR performance.

\begin{table}[H]
    \centering
    \begin{tabular}{@{}lccc@{}}
        \toprule
        \textbf{Configuration} & \textbf{PA-MPJPE} & \textbf{MPJPE} & \textbf{PVE} \\
        \midrule
        Reg-S \small{(HMR2.0\textsuperscript{$\dagger$})} & 41.5 & 67.2 & 81.2 \\
        Reg-M & 39.3 & 65.3 & 78.1 \\
        Def-S & 41.1 & 66.4 & 79.5 \\
        Def-M \small{(\methodname)} & \textbf{\underline{38.3}} & \textbf{\underline{63.6}} & \textbf{\underline{75.2}}\\
        \bottomrule
    \end{tabular}
    \caption{Performance on 3DPW \cite{vonMarcard2018} for $1)$ single-query versus multi-query and $2)$ regular cross-attention versus our proposed deformable cross-attention. Models with "-S" are single-query models, and models with "-M" are multi-query. "Reg" models use decoders with regular cross-attention, while "Def" employ our deformable cross-attention. Reg-S represents our reimplemtation of HMR2.0 with our training data and losses (which we call HMR2.0\textsuperscript{$\dagger$}), and Def-M is \methodname .}
    \label{tab:query_ablation}
\end{table}


\begin{table*}[!htb] 
    \centering
    \begin{tabular}{ccc|ccc|ccc}
        \toprule
        \# \textbf{Attn Heads} & \# \textbf{Groups} ($\mathbf{G}$) & \textbf{Offset Range} ($\mathbf{\lambda_o}$) & \multicolumn{3}{c|}{\textbf{\textit{3DPW}} \cite{vonMarcard2018}} & \multicolumn{3}{c}{\textbf{\textit{RICH}} \cite{huang2022capturinginferringdensefullbody}} \\
        \cmidrule(lr){4-6} \cmidrule(lr){7-9}
        & & & \textbf{PA-MPJPE} & \textbf{MPJPE} & \textbf{PVE} & \textbf{PA-MPJPE} & \textbf{MPJPE} & \textbf{PVE} \\
        \midrule
        16 & 8 & 1 & \textbf{\underline{38.3}} & 63.6 & 75.2 & 48.6 & \textbf{\underline{84.2}} & \textbf{\underline{94.5}} \\
        16  & 8 & 2 & 38.5 & 63.5 & 75.4  & 49.0 & 84.4 & 95.4 \\
        8 & 4 & 1 & 39.2 & 63.2 & 75.5 & 49.0 & 85.8 & 96.2 \\
        8  & 4 & 2 &  38.6 & \textbf{\underline{62.5}} & \textbf{\underline{74.5}} &  \textbf{\underline{48.3}} & 85.3 & 95.8 \\
        \bottomrule
    \end{tabular}
    \caption{Comparison of \methodname\ model performance on 3DPW and RICH datasets across various configurations of deformable cross-attention. We vary the $1)$ number of attention heads, $2)$ number of offset groups, and $3)$ offset range factor. We keep the ratio between the number of attention heads and number offset groups the same in order to scale the information capacity of each offset and head equally.}
    \label{tab:model_comparison}
\end{table*}

\begin{figure}[ht]
    \centering
    \includegraphics[width=0.35\textwidth]{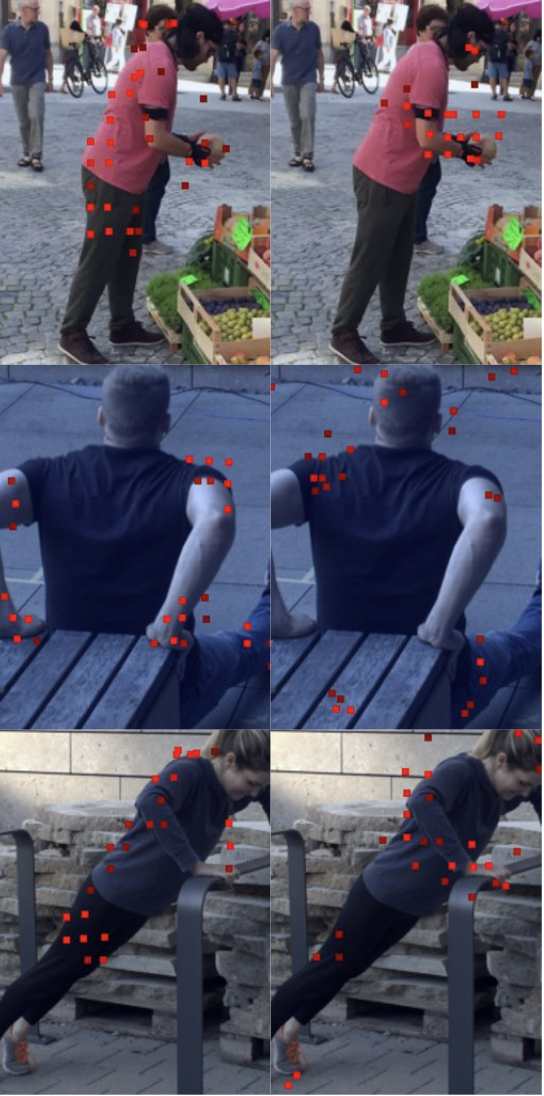}
    \caption{Deformable Attention visualizations for first (\textbf{left}) and last (\textbf{right}) layer of decoder. In each pair of images, we can see important body areas where \methodname\ focuses on in order to model these challenging poses.}
    \label{fig:def_attn_locs}
\end{figure}

In Figure \ref{fig:def_attn_locs}, we visualize what the first attention head of the first (left) and last (right) layer in \methodname\ decoder attends towards. The maroon and red square dots are context positions where the attention value sum over the queries is over 0.25, with bright red corresponding to the largest values. \methodname\ is able to incorporate specific information of each individual's limb positions well through the deformable cross-attention, and we can see that the most important positions correspond well with the attention values and locations. Specifically, in the second and third row the model is able to attend towards uncommon arm and leg positions accurately.

\section{Conclusion}
Through this work, we push the boundaries of HMR by combining a pretrained vision transformer encoder with novel deformable cross attention. We have two main contributions. First, we introduce \methodname, a regression-based framework for monocular HMR that demonstrates SOTA performance on popular 3D HMR datasets such as 3DPW \cite{vonMarcard2018} and RICH \cite{huang2022capturinginferringdensefullbody}. Second, inspired by the self-attention mechanism proposed in the Deformable Attention Transformer (DAT) \cite{xia2023datspatiallydynamicvision}, we extend their method with an innovative deformable cross-attention transformer decoder. This mechanism is both query-agnostic and spatially adaptive, enabling the model to dynamically shift focus on relevant spatial features. We show our decoder performs well without additional encoder fine-tuning, allowing for this method to be applicable for API-based large scale models as well.

Despite these advancements, our work possesses some limitations. While our approach demonstrates significant improvements, there is always room for enhancing the model's robustness, particularly towards examples that are inherently more challenging due occlusions and varying lighting conditions in real-world in-the-wild scenarios. Notably, occlusion from obstacles as well as self-occlusion presents a challenge for the model, particularly noticeable in scenarios where one limb occludes another, such as arms or legs during walking motions. These situations often result in inaccurate limb positioning.

Our work reveals several promising future directions, both within HMR and in other applications. Given the effectiveness of deformable cross-attention for decoding information from spatial features, we believe this method can easily be applied to lower-level tasks such as object detection, instance segmentation, keypoint detection, and pose estimation. Moreover, a potential avenue for advancement is applying our deformable attention towards video data and temporal HMR, dynamically attending towards relevant temporal frames as needed. All things considered, \methodname\ provides a new effective form of decoding spatial features, a paramount necessity in future applications for large pretrained vision models.

\bibliographystyle{ieeenat_fullname}
\bibliography{main}

\clearpage
\setcounter{page}{1}
\maketitlesupplementary

\renewcommand*{\thesection}{\Alph{section}}
\setcounter{section}{0}

\section{Training and Evaluation Details}
We train on two NVIDIA TITAN-RTX GPUs with DDP and global batch size 200. We use AdamW optimizer with learning rate $10^{-4}$ and weight decay $10^{-3}$. For both training and evaluation, in each provided scene, we crop the bounding box of each person and resize it to 256 by 192. 

\section{Additional Ablation Studies}
\subsection{Effect of positional encoding type}
\begin{table}[!htb] 
    \centering
    \begin{tabular}{c|c|c}
        \toprule
        \textbf{PE Type} & \multicolumn{1}{c|}{\textbf{\textit{3DPW}} \cite{vonMarcard2018}} & \multicolumn{1}{c}{\textbf{\textit{RICH}} \cite{huang2022capturinginferringdensefullbody}} \\
        \cmidrule(lr){2-2} \cmidrule(lr){3-3}
        & \textbf{MPJPE} & \textbf{MPJPE} \\
        \midrule
        No PE & 65.4 & 87.0 \\
        Absolute PE & 64.0 & 86.8 \\
        Relative PE & \underline{\textbf{63.6}} & \underline{\textbf{84.2}} \\
        \bottomrule
    \end{tabular}
    \caption{Comparison of \methodname\ model performance on 3DPW and RICH datasets for different positional encoding types. We can observe that the relative positional encoding implementation that we implement results in performance gains, particularly for the out of distribution RICH evaluation dataset.}
    \label{tab:pe_copmparison}
\end{table}


\end{document}